\documentclass[10 pt, conference]{ieeeconf} 
\IEEEoverridecommandlockouts
\pdfoutput=1
\usepackage{cite}
\usepackage{amsmath,amssymb,amsfonts}
\usepackage{algorithmic}
\usepackage{graphicx}
\usepackage{textcomp}
\usepackage{xcolor}
\usepackage{amsmath}
\usepackage{amssymb}

\def\BibTeX{{\rm B\kern-.05em{\sc i\kern-.025em b}\kern-.08em
    T\kern-.1667em\lower.7ex\hbox{E}\kern-.125emX}}
\begin{document}

\title{\LARGE \bf RCM-constrained manipulator trajectory tracking\\using differential kinematics control

\thanks{$^{1}$O.~Rayyan, V.~Gon\c{c}alves and A.~Tzes are with Center for Artificial Intelligence and Robotics (CAIR), New York University Abu Dhabi (NYUAD), 129188, Abu Dhabi, UAE.}%
\thanks{$^{2}$N.~Evangeliou is with Electrical Engineering, NYUAD, UAE.}%
\thanks{$^{2}$Corresponding author's email:
        {\tt\small vmg6973@nyu.edu}.}%

\thanks{Paper presented at the 21st International Conference on Advanced Robotics (ICAR 2023)}
}

\author{Omar Rayyan$^{1}$, Vinicius Gon\c{c}alves$^{1}$, Nikolaos Evangeliou$^{2}$ and Anthony Tzes$^{1}$
}

\maketitle

\begin{abstract}
This paper proposes an approach for controlling surgical robotic systems, while complying with the Remote Center of Motion (RCM) constraint in Robot-Assisted Minimally Invasive Surgery (RA-MIS). In this approach, the RCM-constraint is upheld algorithmically, providing flexibility in the positioning of the insertion point and enabling compatibility with a wide range of general-purpose robots. The paper further investigates the impact of the tool's insertion ratio on the RCM-error, and introduces a manipulability index of the robot which considers the RCM-error that it is used to find a starting configuration. To accurately evaluate the proposed method’s trajectory tracking within an RCM-constrained environment, an electromagnetic tracking system is employed. The results demonstrate the effectiveness of the proposed method in addressing the RCM constraint problem in RA-MIS.
\end{abstract}


\section{Introduction}
Minimally Invasive Surgery (MIS) involves performing an operation through small incisions, with the goal of minimizing tissue disruption. This approach is attributed to reducing patients' trauma, resulting in faster recovery and lower hospitalization costs \cite{fuchs2002minimally}. The advent of Robot-Assisted MIS (RA-MIS) in recent years has further propelled these benefits, allowing surgeons to perform more precise movements with greater agility~\cite{VitielloRev2013,arkenbout2015state, EvangeliouBioRob,SILS_2015}. 

In RA-MIS, the workspace of the robot’s tool-tip is confined to the interior space of the patient's body, which can only be accessed through a designated insertion point. Once the tool-tip of the robot passes through this point, all subsequent movements made by the robot must adhere to the constraints imposed by its location, avoiding any contact with its surrounding boundary. This kinematic constraint resulting from the fulcrum point is widely recognized in the field of RA-MIS as the RCM-constraint. Any deviation from this constraint can lead to an unsafe contact force being exerted on the patient's skin, potentially causing more extensive wounds and longer recovery times. Therefore, minimizing drift from the RCM while maintaining precise tool alignment is of high importance when performing a surgical operation.

The methods employed to maintain the RCM-constraint in RA-MIS are generally categorized in mechanical and software-based approaches. In mechanical-based approaches, the RCM-constraint is maintained inherently through the mechanical structure of the robot. A widely adopted mechanical mechanism makes use of the dual-parallelogram design. In such a system, the RCM-constraint is upheld by ensuring the intersection of the adjacent sides of parallelogram linkages~\cite{kuo2009robotics}. This mechanism is also utilized by the da Vinci surgical system, one of the leading commercialized surgical systems nowadays. On the other hand, software-based approaches control the manipulator while actively ensuring the RCM-constraint is maintained through the utilization of software-based algorithms.

Mechanical approaches, although robust, come with several limitations. Their reliance on the physical design of the robot restricts them into having a single fixed RCM. This results in a significantly reduced adaptability, as it mandates alignment of the insertion point with the RCM prior to each operation. It is also worth mentioning that mechanically constraining a robot to an RCM greatly limits the available space above the insertion point, thereby reducing the range of motion for a robotic arm and obstructing access
~\cite{s23063328}.

Software approaches, 
offer a greater flexibility to surgeons by enabling them to position the RCM directly on any insertion point algorithmically, eliminating the need for an ideal operation table positioning \cite{s23063328}. Moreover it can potentially accommodate non-typical insertion points on the patient’s body and its inherent non-reliance on specialized robot mechanisms, enables the achievement of the RCM-constraint on a wide range of manipulators. 

The main objective of this paper is to propose an enhanced framework for actively controlling a kinematically redundant manipulator while adhering to the RCM-constraint. The proposed control approach makes use of the task function approach and 
operates at a joint-velocity control level. Within the robot-task's function framework
, the tool-tip tracking task is treated as a hard constraint while also ensuring that the task dynamics align with the RCM-constraint, ultimately minimizing any deviation from the fulcrum point. In Subsection~\ref{subs:comparing_appraoches}, comparisons will be done between the presented approach and other approaches also trying to maintain the RCM-constraint algorithmically. Moreover, the accuracy of the proposed method is verified using an external position measurement system rather than using the inherent feedback of the robot.

The core contributions of this paper are: a) Formulation of a minimal task function that considers the RCM-constraint in Subsection (\ref{subs:task_function}), b) Analysis of the tool insertion ratio effect on the RCM-error in Subsection (\ref{subs:tool_insert}), c)  Quantification of a manipulability index which considers the RCM-error in Subsection (\ref{subs:sagsc}), that is then used to generate an initial configuration, and d) Validation of the tool-tip trajectory tracking accuracy through the usage of Aurora electromagnetic tracking system in Subsection (\ref{subs:evaluation}).

\section{Methodology}

\subsection{Mathematical Notation}

All vectors are considered column vectors. For a vector/matrix $M$, $M^{\top}$ represents its transpose. $0_{n \times m}$ and $I_{n \times n}$ represents the $n \times m$ zero matrix and $n$ identity matrix, respectively. $\|\cdot\|$ represents the Euclidean norm of a vector and $\dot{\mathbf{v}}$ the time derivative of a vector $\mathbf{v}(t)$. For two vectors $\mathbf{a}, \mathbf{b}$, the symbol $\mathbf{a} \times \mathbf{b}$ represents their cross product.

\subsection{Forward and Differential Kinematics for the RCM-Constraint}
\label{subs:task_function}
Let $\mathbf{x}_T, \mathbf{y}_T, \mathbf{z}_T$ and $\mathbf{p}_T$ be 3D-vectors representing the $x,y,z$ axis and center, respectively, of a frame $\mathcal{F}_T$ attached to the tool-tip, written in a fixed, world frame $\mathcal{F}_W$. Under the assumption that a cylindrical tool is present, the point $\mathbf{p}_T$ is at the tip of the tool, aligned with the cylinder's axis of rotation axis. The vector $\mathbf{z}_T$ is also aligned with this axis. Subsequently, these vectors can be computed using forward kinematics as a function of the joint position configuration $\mathbf{q}$. Furthermore, let $\mathbf{p}_F$ be the fixed position of the RCM point, also expressed as $\mathcal{F}_W$.

\begin{figure}[htbp]
\includegraphics[width=0.9\columnwidth]{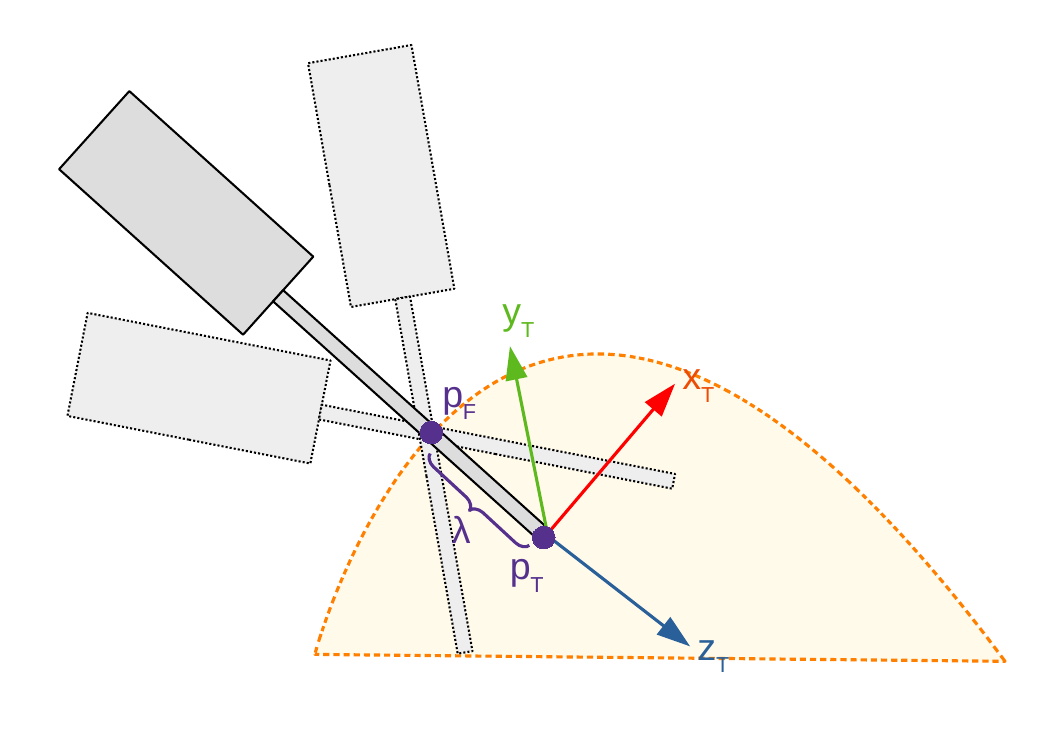}
\caption{Illustration of the RCM-constraint.}
\label{fig:illustfp}
\end{figure}

In Figure~\ref{fig:illustfp} an illustration of the RCM constraint in RA-MIS is depicted. The grey objects represent the end-effector and tool of the robot, in three different configurations. The tool is inserted into the yellow volume through a fulcrum point $\mathbf{p}_F$. The tool-tip position $\mathbf{p}_T$ can be moved as long as one of the points of the tool contains the static point $\mathbf{p}_F$. It can be derived that in order to satisfy the RCM point constraint, there must exist a scalar $\lambda > 0$ such that 
\begin{equation}
\label{eq:fp1}
\mathbf{p}_T(\mathbf{q}) - \mathbf{p}_F = \lambda \mathbf{z}_T(\mathbf{q}).
\end{equation}
The scalar $\lambda$ represents how much the tool is inserted into the RCM, and is positive if and only if the tool is inserted. Pre-multiplying this equation by $\mathbf{x}_T^{\top}$ and then by $\mathbf{y}_T^{\top}$ and using the orthogonality condition on $\mathbf{x}_T, \mathbf{y}_T, \mathbf{z}_T$, \eqref{eq:fp2} holds.
\begin{eqnarray}
\label{eq:fp2}
  &&  \mathbf{x}_T(\mathbf{q})^{\top}\big(  \mathbf{p}_T(\mathbf{q}) - \mathbf{p}_F ) = 0, \nonumber \\
  &&  \mathbf{y}_T(\mathbf{q})^{\top}\big(  \mathbf{p}_T(\mathbf{q}) - \mathbf{p}_F ) = 0.
\end{eqnarray}

The benefit of using  \eqref{eq:fp2} over \eqref{eq:fp1} is that the latter does not have the variable $\lambda$. Clearly, \eqref{eq:fp2} can be written  as $\mathbf{r}_{F}(\mathbf{q})=0_{2 \times 1} $, for a function $\mathbf{r}_{F} : \mathbb{R}^n \mapsto \mathbb{R}^2$. This is the \emph{task function} for the RCM-constraint that will be used for controlling the robot. Note that the RCM-constraint requires a robot configuration with a minimum of two degrees of freedom, one for translational movements along the insertion axis and another for rotating the insertion axis alongside the fulcrum point. These two degrees of freedom align with the number of components of the proposed task function.

In order to implement the controller, the task Jacobian  will be expressed as $J_F(\mathbf{q}) \triangleq \frac{\partial \mathbf{r}_F}{\partial \mathbf{q}}(\mathbf{q})$. The first row of this task Jacobian, relative to the first component of $\mathbf{r}_F$,  will be derived, and the second one comes using a very similar reasoning. For this, let $J_v(\mathbf{q})$ and $J_{\omega}(\mathbf{q})$ be the linear and angular velocity Jacobians, respectively, for the tooltip frame $\mathcal{F}_T$ written in $\mathcal{F}_W$. From standard differential kinematics~ \cite{siciliano2010robotics} it is known that 
\begin{itemize}
    \item (i) $\frac{d}{dt} \mathbf{p}_T(\mathbf{q}) = J_v(\mathbf{q})\dot{\mathbf{q}}$
    .
    \item (ii) Let $\boldsymbol{\omega}$ be the angular velocity of the tooltip frame.  It is known that  $\boldsymbol{\omega}(\mathbf{q},\dot{\mathbf{q}}) = J_{\omega}(\mathbf{q})\dot{\mathbf{q}}$. Furthermore, it is also known that $\frac{d}{dt} \mathbf{x}_T(\mathbf{q}) =\boldsymbol{\omega}(\mathbf{q},\dot{\mathbf{q}}) \times \mathbf{x}_T(\mathbf{q})$. Consequently,  $\frac{d}{dt} \mathbf{x}_T(\mathbf{q}) =  \big( J_{\omega}(\mathbf{q}) \dot{\mathbf{q}} \big) \times \mathbf{x}_T(\mathbf{q})$. 
\end{itemize}.
Furthermore, let $r_{T,x}(\mathbf{q}) \triangleq \mathbf{x}_T(\mathbf{q})^{\top}\big(  \mathbf{p}_T(\mathbf{q}) - \mathbf{p}_F )$ be the first entry of $\mathbf{r}_T$. On one hand, by the chain rule 
\begin{equation}
\label{eq:jacrF1}
    \frac{d}{dt} r_{T,x} = \frac{\partial r_{T,x}}{\partial \mathbf{q}} \dot{\mathbf{q}}.
\end{equation}
On the other hand, by differentiating the expression of $r_{T,x}(\mathbf{q})$ using the results (i) and (ii), \ref{eq:jacrF2} can be derived
\begin{equation}
\label{eq:jacrF2}
     \frac{d}{dt} r_{T,x} = \Big( \big( J_{\omega}\dot{\mathbf{q}} \big) \times \mathbf{x}_T \Big)^{\top}\big(\mathbf{p}_T-\mathbf{p}_F\big) + \mathbf{x}_T^{\top}J_v\dot{\mathbf{q}},
\end{equation}

\noindent in which the dependencies in $\mathbf{x}_T, \mathbf{y}_T, \mathbf{p}_T, J_v$ and $J_w$ on $\mathbf{q}$ are omitted. Now, the mixed product formula is used: $(\mathbf{a} \times \mathbf{b})^{\top} \mathbf{c} =  ( \mathbf{b} \times \mathbf{c})^{\top} \mathbf{a}$. Applying this result into \eqref{eq:jacrF2} and factoring out $\dot{\mathbf{q}}$, ~\ref{eq:jacrF3} can be formulated
\begin{equation}
\label{eq:jacrF3}
     \frac{d}{dt} r_{T,x} = \Big( \mathbf{x}_T^{\top}J_v + \big( \mathbf{x}_T \times (\mathbf{p}_T - \mathbf{p}_F) \big)^{\top} J_\omega \Big)\dot{\mathbf{q}}.
\end{equation}

Comparing \eqref{eq:jacrF1} with \eqref{eq:jacrF3}, 
then $\frac{\partial r_{T,x}}{\partial \mathbf{q}} = \mathbf{x}_T^{\top}J_v + \big( \mathbf{x}_T \times (\mathbf{p}_T - \mathbf{p}_F) \big)^{\top} J_\omega$. Similarly, the Jacobian for the component $r_{T,y}$ is obtained, by replacing $\mathbf{x}_T$ with $\mathbf{y}_T$ in the formula. Stacking these two Jacobians the desired Jacobian for $J_F(\mathbf{q})$ can be obtained. Equation~\ref{eq:Summary} summarizes the above formulations.
\begin{eqnarray}
\label{eq:Summary}
   && \mathbf{r}_F = \left[\begin{array}{c} \mathbf{x}_T^{\top}\big(  \mathbf{p}_T - \mathbf{p}_F ) \\
   \mathbf{y}_T^{\top}\big(  \mathbf{p}_T - \mathbf{p}_F ) \end{array}\right]. \nonumber \\
   && J_F = \left[\begin{array}{c}
    \mathbf{x}_T^{\top}J_v + \big( \mathbf{x}_T \times (\mathbf{p}_T - \mathbf{p}_F) \big)^{\top} J_\omega \\
    \mathbf{y}_T^{\top}J_v + \big( \mathbf{y}_T \times (\mathbf{p}_T - \mathbf{p}_F) \big)^{\top} J_\omega \end{array}\right].
\end{eqnarray}
\subsection{Controller Design}
The controller design is based on the task to track a tool-tip trajectory $\mathbf{p}_{T,d}(t)$, while  keeping the RCM error to $0$, $\mathbf{r}_F = 0_{2 \times 1}$. For that, the \emph{task function approach} \cite{samson1991robot} is incorporated. 
In this approach, it is assumed that the control objective can be codified with  $m$ differentiable functions $\mathbf{r}_i: \mathbb{R}^n \times \mathbb{R} \mapsto \mathbb{R}^{k_i}$ of the configuration $\mathbf{q}$ and time $t$ as $\mathbf{r}_i(\mathbf{q},t) = 0_{k_i \times 1}$ for $i=1,2,...,m$, i.e, the control objective is translated into zeroing all the functions $\mathbf{r}_i$. Then, a \emph{task dynamics specification} is imposed for the purpose of choosing the control inputs so that $\frac{d}{dt} \mathbf{r}_i = -K_i \mathbf{r}_i$ holds along the system trajectories. This case implies that all of the $r_i \rightarrow 0$ exponentially. Therefore, in order to use this approach it is necessary to (i) choose the functions $\mathbf{r}_i$ according to the task, (ii) select the gains $K_i$, that controls how fast the functions $\mathbf{r}_i$ converge to $0$ and how aggressive the controllers are and (iii) select the control inputs to meet the equation $\frac{d}{dt} \mathbf{r}_i = -K_i\mathbf{r}_i$.

For the step (i), for the first task, the task function $\mathbf{r}_{T}(\mathbf{q},t) \triangleq \mathbf{p}_T(\mathbf{q}) - \mathbf{p}_{T,d}(t)$ will be used. For the second task,  the function $\mathbf{r}_F$ defined on the previous subsection will be used. The choice for step (ii) is discussed in Section~\ref{sec:expresults}. Finally, in step (iii)  the control input must be chosen. Using the chain rule and the assumption that $\dot{\mathbf{q}} = \mathbf{u}$, the task dynamics specifications $\frac{d}{dt} \mathbf{r}_i = -K_i\mathbf{r}_i$ become 
\begin{eqnarray}
\label{eq:taskeq}
    && J_v(\mathbf{q})\mathbf{u} = -K_T \mathbf{r}_T(\mathbf{q},t) + \dot{\mathbf{p}}_{T,d}(t) \nonumber \\
    && J_F(\mathbf{q})\mathbf{u} =  -K_F \mathbf{r}_F(\mathbf{q})
\end{eqnarray}

\noindent in which the chain rule was used together with  the fact that $\frac{\partial}{\partial \mathbf{q}} \mathbf{r}_T = J_v$ as well as that $\frac{\partial}{ \partial t} \mathbf{r}_T = -\dot{\mathbf{p}}_{T,d}$.

In this work, since the motion $\mathbf{p}_{T,d}(t)$ will be relatively slow, this feedforward term can be omitted. So, once this term is removed, \eqref{eq:taskeq} becomes a system of linear equation for the variable $\dot{\mathbf{q}} = \mathbf{u}$. Considering both matricial equations for the two tasks $3+2=5$ equations are extracted, whereas for the Kuka LBR IIWA robot $n=7$ joints are present. So, assuming that all the rows of the matrices $J_v$ and $J_F$ are linearly independent, there exist $7-5 = 2$ additional Degrees of Freedom (DoF) and there are infinite solutions for~\eqref{eq:taskeq}. However, it may be the case that some rows of $J_v$ are linearly dependent, or very close to be linearly dependent to the rows of $J_F$. This means that there is a conflict between executing the task dynamic specifications for following the tooltip trajectory and for keeping the RCM-constraint.

In order to solve both issues (infinite solutions and conflicts), the problem of computing $\mathbf{u} = \mathbf{\dot{q}}$ will be formulated as the quadratic program
\begin{eqnarray}
\label{eq:formulation}
    \mathbf{u}(\mathbf{q},t) &=& \arg \min_{\boldsymbol{\mu}} \|J_F(\mathbf{q})\boldsymbol{\mu}+K_F \mathbf{r}_F(\mathbf{q})\|^2 + \epsilon \|\boldsymbol{\mu}\|^2
    \nonumber\\
    && \mbox{such that \ \ } J_v(\mathbf{q})\boldsymbol{\mu} = -K_T \mathbf{r}_T(\mathbf{q},t)
\end{eqnarray}

\noindent in which $\epsilon$ is a small positive constant (in this work, $\epsilon=10^{-6}$). This formulation solves both the infinite solution problem and the conflict problem. Essentially, the program formulates the problem as follows: it treats the tooltip tracking task as a hard constraint, since the respective task dynamics specification is a constraint, and then, considering this, it tries to both satisfy the task dynamics specification for the RCM-constraint (trying to minimize $\|J_F(\mathbf{q})\boldsymbol{\mu}+K_F\mathbf{r}_F(\mathbf{q})\|^2$) and maintaining a low velocity (trying to minimize $\|\boldsymbol{\mu}\|^2)$. However, since $\epsilon$ is small, it prioritizes more having a better control of the RCM than having a small norm. These characteristics guarantee that the solution is unique (trying to minimize $\|\boldsymbol{\mu}\|^2$ guarantees strict convexity of the formulation and consequently uniqueness of solution) and that if there is a conflict between the tooltip tracking task and the RCM task, the former should have more priority. Furthermore, since \eqref{eq:formulation} is a quadratic program without inequality constraints, it can be solved analytically by solving the system
\begin{equation}
\hspace*{-5mm}
\left[\begin{array}{cc} J_F^{\top}J_F+\epsilon I_{n \times n} &  J_v^{\top} \\
  J_v & 0_{3 \times n} \end{array}\right] \left[\begin{array}{c} \mathbf{u} \\ \boldsymbol \gamma \end{array}\right] =  \left[\begin{array}{c} -K_F J_F^{\top} \mathbf{r}_F \\ -K_T \mathbf{r}_T \end{array}\right] 
\end{equation}
\noindent in which $\boldsymbol \gamma$ is the Lagrange multiplier.
\subsection{Comparison with Other Approaches}
\label{subs:comparing_appraoches}
In this section, the presented approach is compared with others that also use the task function framework to formulate the problem. The focus will be on the description of the RCM task only. To make the comparison easier, 
their task functions are rewritten into equivalent forms
.

In \cite{art2010}, the formulation of the RCM-constraint is based on the tangent plane equation at the insertion point, and a task function is derived based on this plane. The selected task function is equivalent to the task function $\mathbf{r}_{F,1}(\mathbf{q}) \triangleq \lambda \mathbf{z}_T(\mathbf{q}) - (\mathbf{p}_T(\mathbf{q}) - \mathbf{p}_F)$. Note that having this task function vanish is equivalent to having \eqref{eq:fp1}. However, it is assumed that $\lambda$ is constant, which is not the case in the proposed approach. 

In \cite{s23063328}, the RCM-constraint task is defined to be 
    $\mathbf{r}_{F,2}(\mathbf{q}) \triangleq (I - \mathbf{z}_T\mathbf{z}_T^T)(\mathbf{p}_T(\mathbf{q})-\mathbf{p}_F(\mathbf{q}))$. This task function incorporates an extra task component when compared to the proposed task function which includes 2 components, matching the DoF required by the RCM-constraint. This means the task function of \cite{s23063328} exhibits a redundant task component. The proposed task function $\mathbf{r}_F$ is related to this task function $\mathbf{r}_{F,2}$ by the expression $\mathbf{r}_F = [\mathbf{x}_T^{\top}\mathbf{r}_{F,2} \ \mathbf{y}_T^{\top}\mathbf{r}_{F,2}]^{\top}$.

In \cite{art2019}, the RCM-constraint task is defined to be
  $\mathbf{r}_{F,3}(\mathbf{q,\lambda}) \triangleq \lambda \mathbf{z}_T(\mathbf{q}) - (\mathbf{p}_T(\mathbf{q}) - \mathbf{p}_F)$. This task function is equivalent to $\mathbf{r}_{F,1}$, except that $\lambda$ is no longer constant. This requires $\lambda$ to be continuously evolved according to the feedback in the closed loop. In addition to that, this approach requires the usage of a transformation matrix to partition the joint velocities, which is not necessary in the proposed approach. 
  

\subsection{The Effect of the Tool Insertion}
\label{subs:tool_insert}
From \eqref{eq:fp1},  pre-multiplying by $\mathbf{z}_T$ to obtain that the \emph{tool insertion} $\lambda$ can be given as $\lambda(\mathbf{q}) = \mathbf{z}_T(\mathbf{q})^{\top}\big(\mathbf{p}_T(\mathbf{q})-\mathbf{p}_F\big)$. Let $L$ be the length of the tool. An important index is the ratio between how much is inserted into the RCM, $\lambda$ and how much is outside, $L-\lambda$. This ratio, $\rho(\mathbf{q}) \triangleq |(L-\lambda(\mathbf{q}))/\lambda(\mathbf{q})|$ is important. Essentially, the higher is this number, the more difficult it is to control the robot because a small movement in the tool-tip may require a large movement for the robot as whole.

\begin{figure}[h]
\centering
\includegraphics[width=0.8\columnwidth]{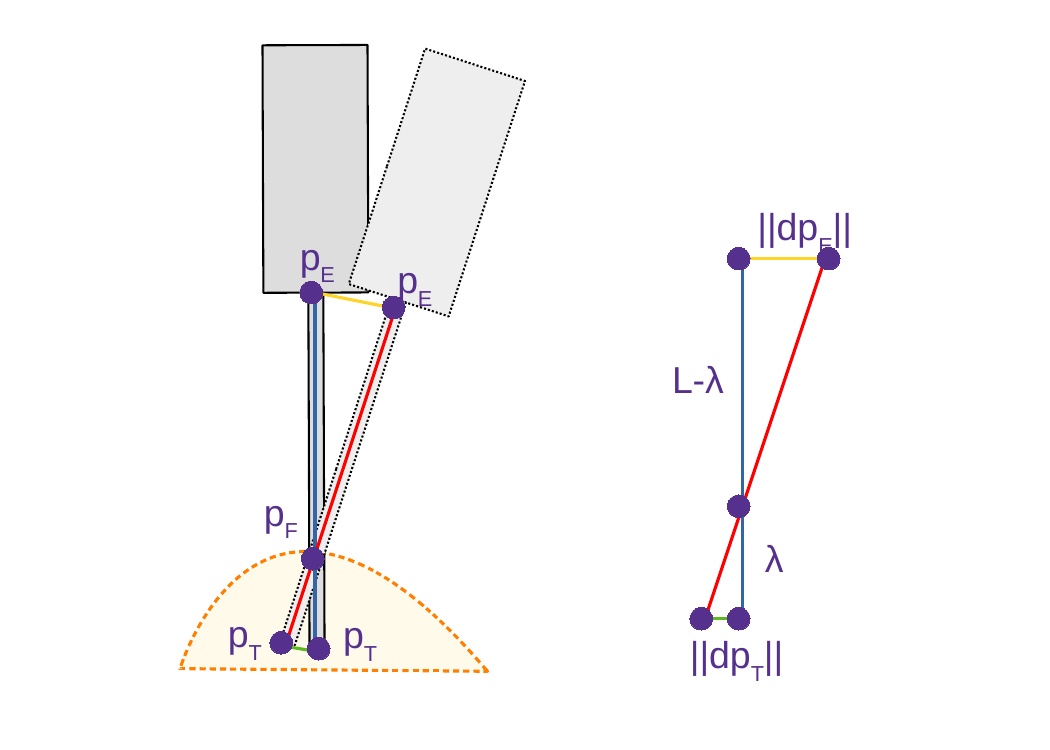}
\caption{Geometric illustration of the fact that $\|d\mathbf{p}_E\|/\|d\mathbf{p}_T\| = (L-\lambda)/\lambda = \rho$. If $\|d\mathbf{p}_E\|$ and  $\|d\mathbf{p}_T\|$ are small, the two triangles formed are very close to be right triangles, similar to each other, and thus, $\|d\mathbf{p}_E\|/(L-\lambda) = \|d\mathbf{p}_T\|/\lambda$
.}
\label{fig:rhoillust}
\end{figure}

This is better visualized by considering how a small change in the tooltip position $\|d\mathbf{p}_T\|$ is related to a small change in a point $\mathbf{p}_E$ that connects the tool to the robot, that is, $\|d\mathbf{p}_E\|$, as shown in Figure~\ref{fig:rhoillust}. Since the interest is on studying how the insertion affects the movement, it is assumed that in this movement the insertion does not change, so $d\lambda = 0$. From \eqref{eq:fp1}, one can take the differential from both sides, use the fact that $d\mathbf{p}_F=0$ (since the RCM is constant) and $d\lambda =0$ (it is assumed that the insertion does not change) to obtain:
\begin{equation}
\label{eq:dpT1}
    d\mathbf{p}_T = \lambda d\mathbf{z}_T
\end{equation}

\noindent and thus $\|d\mathbf{p}_T\| = | \lambda| \|d\mathbf{z}_T\|$. Furthermore, since it is possible to write $\mathbf{p}_T = \mathbf{p}_E + L \mathbf{z}_T$, taking the differential from both sides 

\begin{equation}
\label{eq:dpT2}
    d\mathbf{p}_T = d\mathbf{p}_E + L d\mathbf{z}_T
\end{equation}

\noindent and thus from \eqref{eq:dpT1} and \eqref{eq:dpT2} the result $ d\mathbf{p}_E = (\lambda-L) d\mathbf{z}_T$ is obtained and, consequently, $\|d\mathbf{p}_E\| = |\lambda-L| \|d\mathbf{z}_T\|$. Therefore, $\|d\mathbf{p}_E\|/\|d\mathbf{p}_T\| = |\lambda-L|/|\lambda| = \rho$. This result means that $\rho$ measures, in a given configuration, how much a (small) movement in the tooltip generates of movement in a specific point of the robot (the one that connects the robot to the tool) or, equivalently, how the respective velocities relate. 
So, for example, $\rho = 5$ means that having $\|\dot{\mathbf{p}}_T\| = 1$cm/s on the tooltip will require, roughly, that $\mathbf{p}_E$ moves with velocity $\|\dot{\mathbf{p}}_E\| = 5$cm/s, i.e, small velocities on the tooltip generate large velocities on the robot which, then, requires large velocities on the joints. Indeed, as it will be shown in the experimental results, generally when $\rho$ increases the RCM error $\mathbf{r}_F$ also increases. This implies that, as a general rule, the tool should be designed so $\rho(\mathbf{q})$ is as small as possible. 
\subsection{Selecting the Starting Configuration}
\label{subs:sagsc}
A crucial parameter in the surgical room is the positioning of the robotic equipment with respect to the patient's body and the surrounding physicians. For every operation there always exists an optimal initial joint space configuration for a robotic arm that allows for optimal workspace coverage and dexterity. 

This issue can be quantified using the \emph{manipulability} index. Suppose that, at a configuration $\mathbf{q}_0$ and with a RCM $\mathbf{p}_F$, it is desired to execute a linear velocity $\mathbf{v}$ on the tooltip. It is assumed that the RCM error is zero at $\mathbf{q}_0$. So, the joint velocity $\dot{\mathbf{q}}$ can be decided by solving the system of equations $J_F(\mathbf{q}_0, \mathbf{p}_F)\dot{\mathbf{q}} = 0_{2 \times 1}$ (keep the RCM error in $0$) and $J_v(\mathbf{q}_0)\dot{\mathbf{q}} = \mathbf{v}$ (move according to this linear velocity). Note that it is written $J_F(\mathbf{q}_0, \mathbf{p}_F)$ to stress that the RCM Jacobian also depends on the RCM $\mathbf{p}_F$. Let $J(\mathbf{q}_0,\mathbf{p}_F) = J \triangleq  [J_v^{\top} \ J_F^{\top}]^{\top}$. Then, it is not difficult to show that the smallest solution $\dot{\mathbf{q}}$ (in the Euclidean norm) to these two equations has Euclidean norm $\|\dot{\mathbf{q}}\| = \sqrt{\mathbf{v}^T M \mathbf{v}}$, in which $M(\mathbf{q}_0,\mathbf{p}_F)$ is the top $3$ rows and columns of the matrix $(JJ^{\top})^{-1}$.  So, if the eigenvalues of the matrix $M(\mathbf{q}_0,\mathbf{p}_F)$ are large, this means that in order to execute a linear velocity at that configuration $\mathbf{q}_0$ and RCM $\mathbf{p}_F$,  large joint velocities are required. So, it is reasonable to choose a configuration $\mathbf{q}_0$ and RCM $\mathbf{p}_F$ to operate around so $M$ has small eigenvalues.

It is necessary, therefore, to optimize both on $\mathbf{q}_0$ and $\mathbf{p}_F$, In order to simplify the problem,  the RCM choice will be attached to the choice of $\mathbf{q}_0$ as follows: $\mathbf{p}_F = \mathbf{p}_T(\mathbf{q}_0) - \lambda_0  \mathbf{z}_T(\mathbf{q}_0)$, in which $\lambda_0=0.1m$ is fixed. This means that the RCM is selected as being $10cm$ above the tooltip along the tool at the starting configuration. With this, one can optimize only on $\mathbf{q}_0$, by minimizing the maximum eigenvalue (spectral radius) of $M$ while considering other constraints as the tooltip position being inside a region in the workspace, the configuration being within the joint limits and also that the configuration should make the tool be pointing downwards. 
\section{Experimental Studies}
\label{sec:expresults}

\subsection{Experimental Setup}
\label{subs:exp_setup}

\begin{figure}[htbp]
    \centering
    \includegraphics[width=0.9\linewidth]{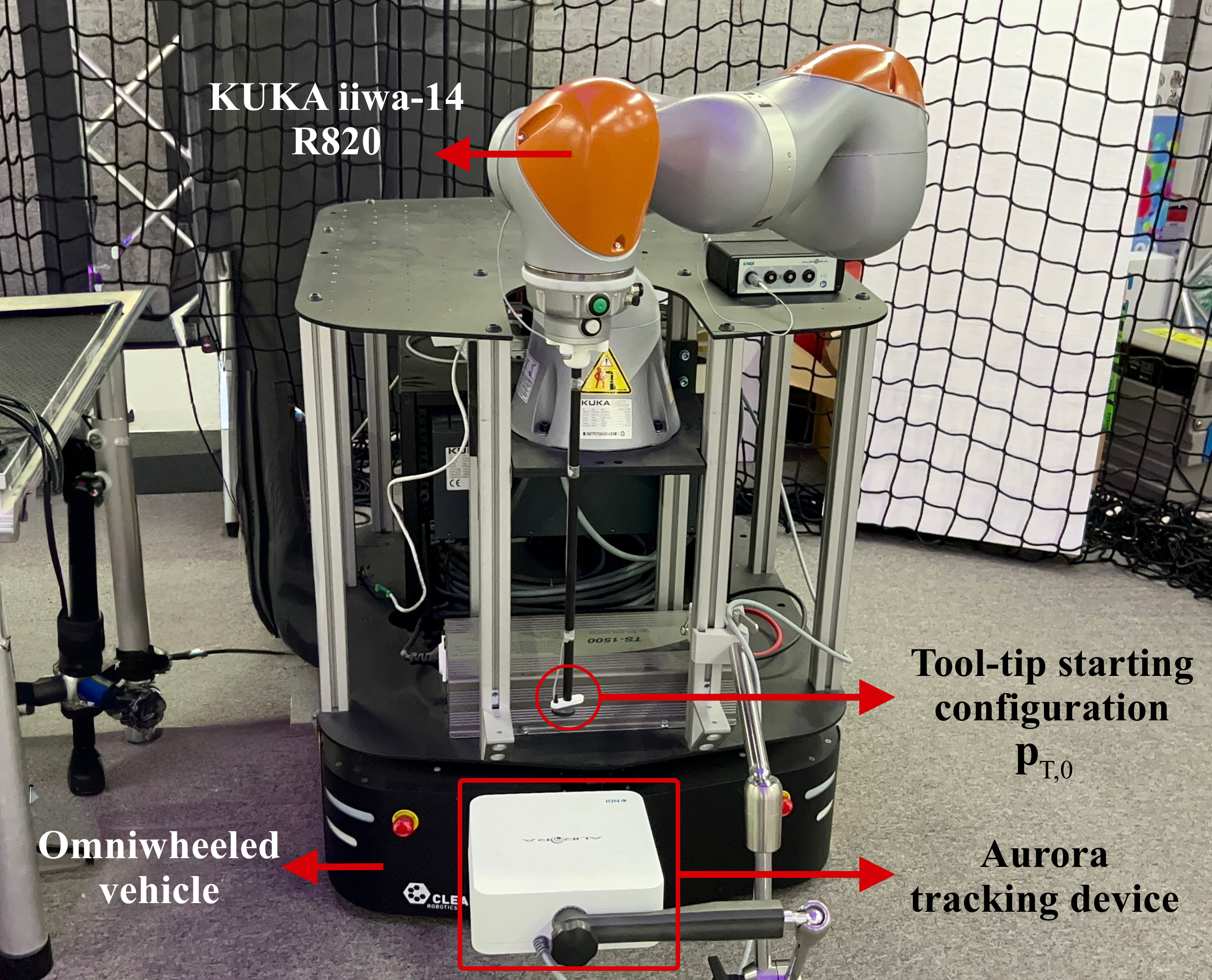}
    \caption{Initial Configuration of the Manipulator ($\mathbf{p}_{T,0}$) in the setup environment}
\label{fig:initial_config}
\end{figure}

To verify the accuracy of the proposed controller, it was tested  on a 7-DOF KUKA LBR iiwa-14 R820 robotic manipulator, shown in Figure~\ref{fig:initial_config} using KUKA's Fast Research Interface (FRI) \cite{iiwa2019github} which offers joint-positions control. Considering the imposed constraint in FRI of manipulating the serial robotic arm through joint-positions, and given that the proposed controller operates on the joint-velocity level, a virtual joint-velocity control method was employed. After determining the joint-velocities $\mathbf{\dot{q}_{n}}$, numerical integration is performed using the preceding joint-positions $\mathbf{q_{n-1}}$ to compute the subsequent joint-positions  $\mathbf{q_{n}}$, applying the formula $\mathbf{q_{n}=q_{n-1}+\frac{\dot{q}_{n}}{frequency}}$. This allows the controller to be adapted to work with the KUKA's FRI joint-position control interface at a frequency of 250 Hz.

In order to mimic the surgical instruments used in RA-MIS procedures, a tool was attached to the end-effector of the manipulator, having a length of $L=0.4m$. The RCM constraint was then established at a distance of $\lambda_0=0.1m$  from the tool-tip. As a result, an initial insertion ratio of $\rho(\mathbf{q_0}) = \frac{0.3m}{0.1m} = 3 $ was established. To generate an initial configuration that maximizes manipulability, the annealing algorithm, described in Subsection \ref{subs:sagsc}, was used to get $\mathbf{q}_{0}=[35.5^\circ,81.9^\circ,-92.2^\circ,-92.0^\circ,82.1^\circ,91.2^\circ,-72.0^\circ]^{\top}$, which corresponds to a starting position $\mathbf{p}_{T,0}=[0.562m,-0.095m,-0.126m]$. This starting configuration can be seen in Figure~\ref{fig:initial_config}. The fulcrum point $\mathbf{p}_F$ is then established to be at a distance $\lambda=\frac{L}{1+\rho(q_0)}$ from tool-tip of this starting configuration. The robot was then instructed to trace the helical trajectory described by the parametric equation: 
\begin{equation}
\label{eq:traj}
\mathbf{p}_{T,d}(t) = \mathbf{p}_{T,0} +\begin{bmatrix} 0.03\alpha(t)\cos{(\frac{\pi}{5}t)}\\ 0.03\sin{(\frac{\pi}{5}t)}\\ 0.06\sin{(\frac{\pi}{10}t)}-0.04\alpha(t)
\end{bmatrix}
\end{equation}
\noindent in which $\alpha(t)=\max(1,t/5)$, $t$ is in seconds and $\mathbf{p}_{T,d}(t)$ in meters. The function $\alpha(t)$ was incorporated in the helical path parametric equation to ensure a smooth transition from the initial tooltip position to the initial desired tooltip position (so  $\mathbf{p}_{T,d}(0)=\mathbf{p}_{T,0}$). The helical path was chosen as it serves as an ideal approximation of suturing motion - a very common task in surgical procedures~\cite{karadimos2022perception}. The parameters $K_T$ and $K_F$ in \eqref{eq:taskeq} were set to  $K_T=14 s^{-1}$ and $K_F=27 s^{-1}$.
\subsection{Results and Accuracy Evaluation}
\label{subs:evaluation}
\begin{figure}[htbp]
    \centering
    \includegraphics[width=0.9\linewidth]{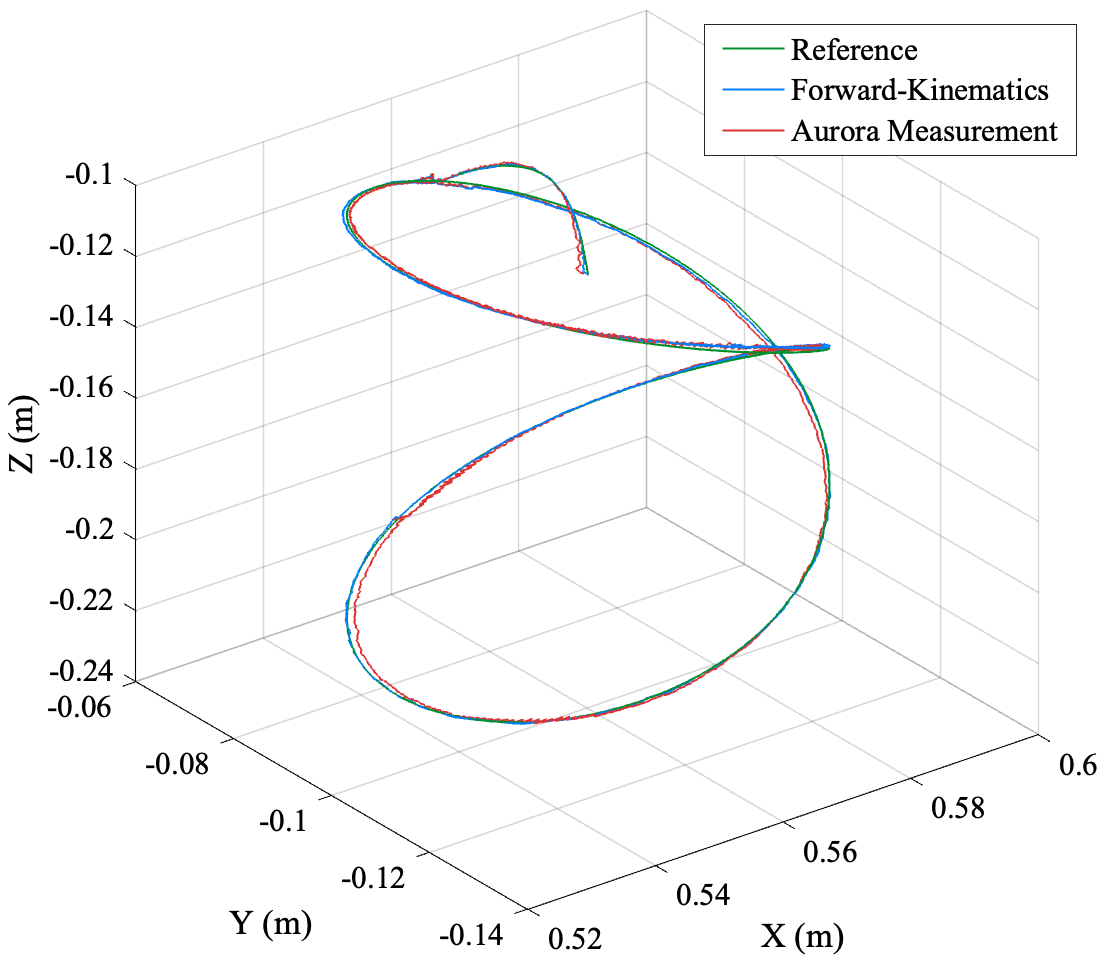}
    \caption{Tool-Tip Position Visualization using the forward-kinematics model and the Aurora device measurement}
    \label{fig:tooltip}
\end{figure}

\begin{figure}[htbp]
    \centering
    \includegraphics[width=0.9\linewidth]
    {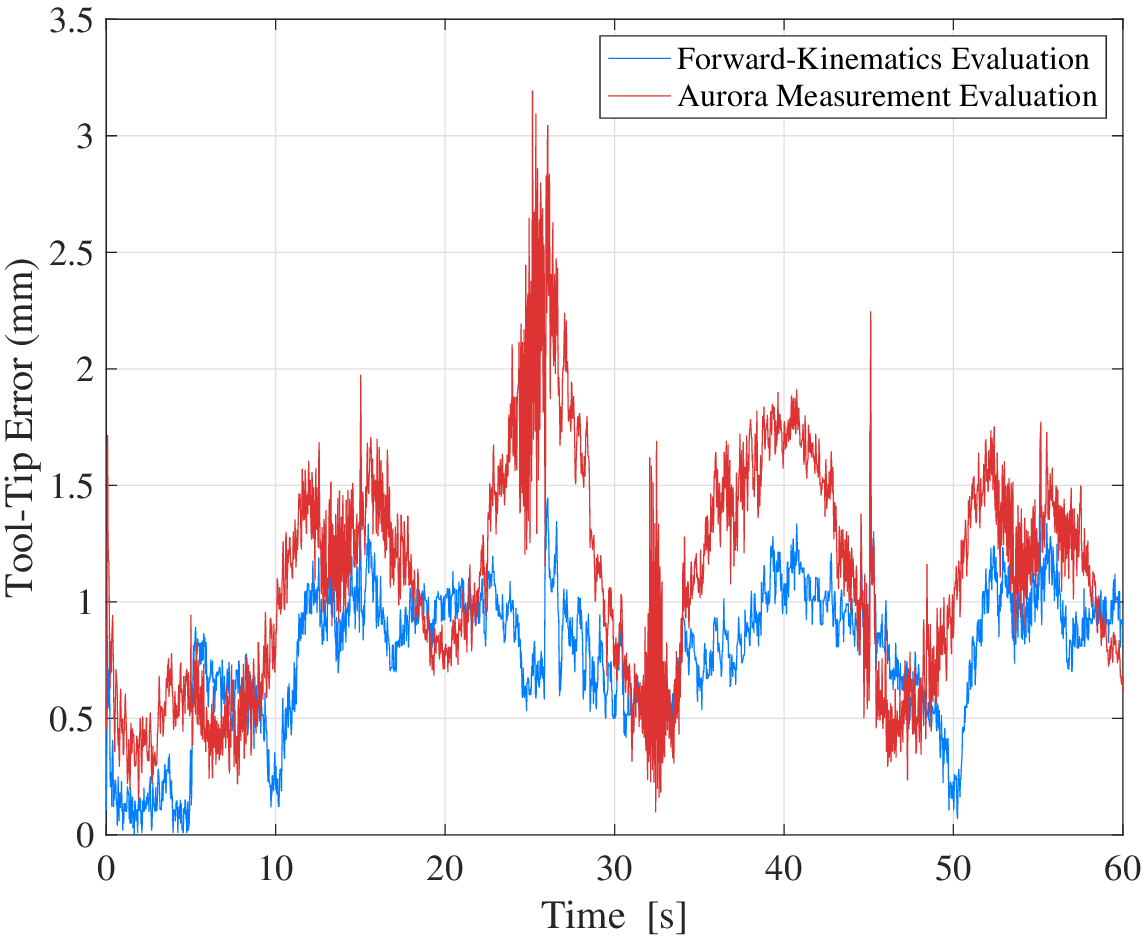}
    \caption{Tool-Tip Position Error evaluated using the forward-kinematics model and the Aurora device measurement}
    \label{fig:positional_error}
\end{figure}

To evaluate the accuracy of the tool-tip tracking along the helical path in \eqref{eq:traj}, two distinct measurement methods were used to continuously determine the tool-tip’s position. The primary method involves using the manipulator's joint position feedback for calculating the tip position via the forward kinematics formulation. The secondary method involves the use of an external tracking device to assess the accuracy of the forward kinematics estimation of the first method \cite{hummel2002evaluation,hummel2006evaluation}. Specifically, a version of the Aurora electromagnetic tracking system from NDI with RMS positioning accuracy of $0.8~\textrm{mm}$ 
 and measurement frequency of $40Hz$ was used. 
The tracking of the tool-tip, evaluated using the two layers of accuracy measurements 
, is visualized in Figure \ref{fig:tooltip}. Additionally, Figure~\ref{fig:positional_error} illustrates the positioning error during helical path execution, calculated as  $\|\mathbf{r}_T(\mathbf{q}(t),t)\|$. On average, when calculated using the forward-kinematics model estimation, the positioning error relative to the desired trajectory (\ref{eq:traj}) amounts to $0.78~\textrm{mm}$. However, when evaluated using the Aurora device measurements, the positioning error averages $1.1~\textrm{mm}$ along the path.

\begin{figure}[htbp]
    \centering
    \includegraphics[width=0.9\linewidth]{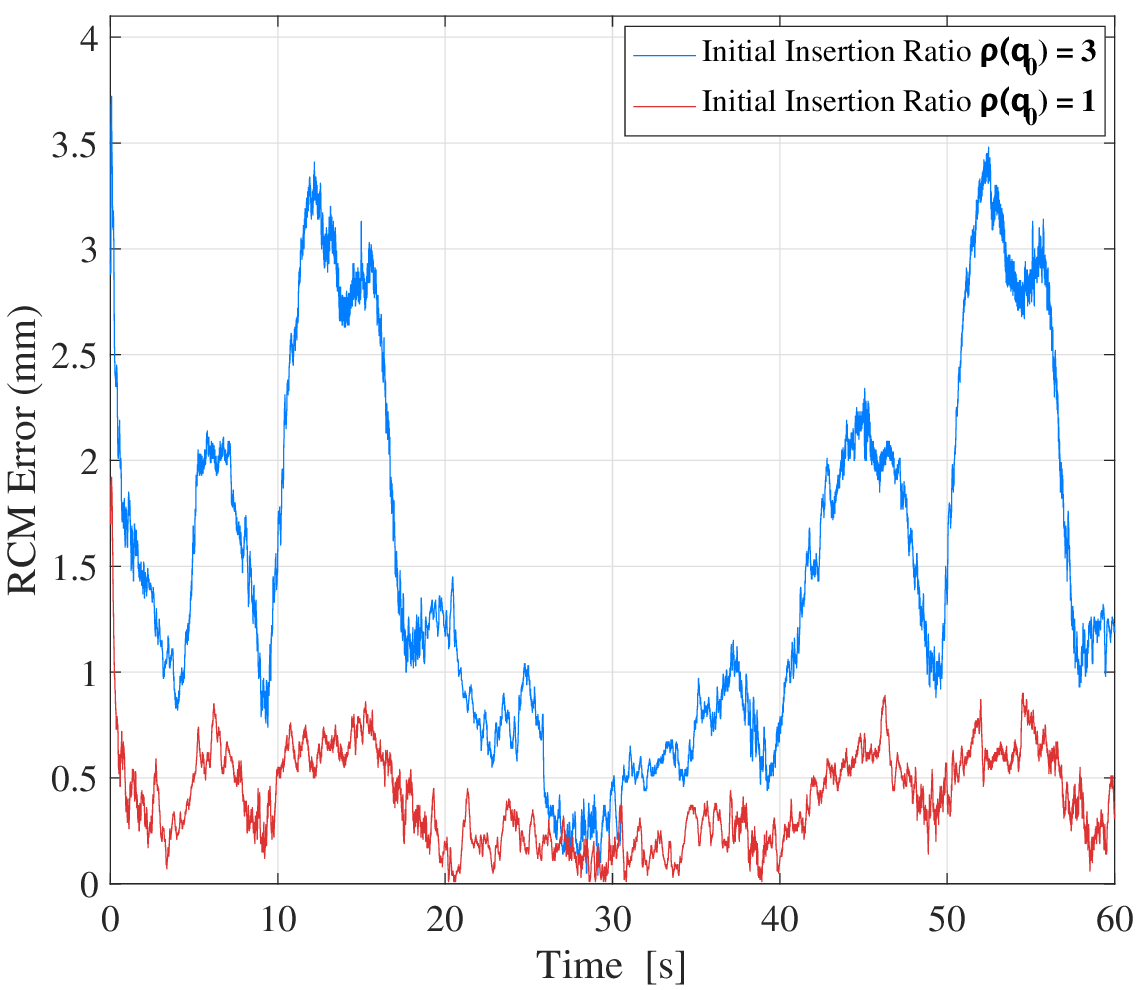}
    \caption{RCM position error $\|\mathbf{r}_F(\mathbf{q}(t))\|$ at an initial insertion ratio of $\rho(\mathbf{q}_0)=3$ and $\rho(\mathbf{q}_0)=1$ evaluated using the forward-kinematics model}
    \label{fig:RCM}
\end{figure}

Figure \ref{fig:RCM} shows the RCM error throughout the tracked trajectory, calculated as $\|\mathbf{r}_F(\mathbf{q}(t))\|$  and equal to the Euclidean distance between the tool and the fulcrum point $\mathbf{p}_F$. With an initial insertion ratio $\rho(\mathbf{q_0}) = 3 $, the RCM following had an average error of $1.5~\textrm{mm}$. To verify the proposed hypothesis in Subsection \ref{subs:tool_insert}, the initial RCM position was later modified to become a distance of $\lambda_0=0.2m$ from the tool-tip. As a result, the initial insertion ratio became $\rho(\mathbf{q_0}) = \frac{0.2m}{0.2m} = 1 $. The error of the manipulator's RCM from the fulcrum point at this insertion ratio is depicted in Figure \ref{fig:RCM}. On average, the RCM error at $\rho(\mathbf{q_0}) = 1 $ was $0.4~\textrm{mm}$, a 73.3\% decrease from that at $\rho(\mathbf{q_0}) = 3$. 

\begin{figure}[htbp]
    \centering
    \includegraphics[width=8cm]{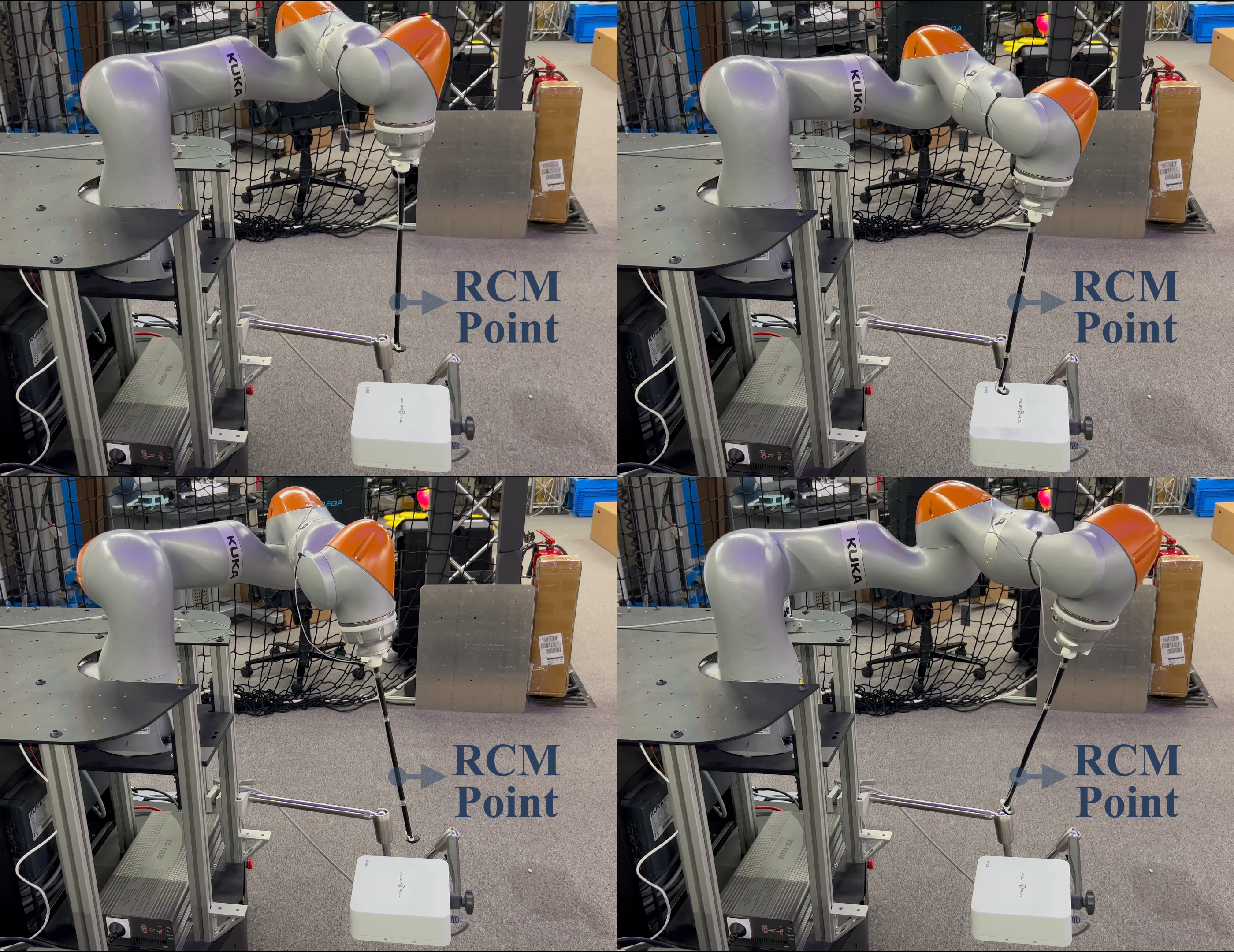}
    \caption{RCM Visualization}
    \label{fig:RCM_Visualization}
\end{figure}

A visualization of random  achieved manipulator configurations during RCM validation is depicted in Figure ~\ref{fig:RCM_Visualization}. Finally, a video of the experiment can be seen in \texttt{https://youtu.be/Ybg34tYok9U} and the relevant open-sourced implementation is located at \texttt{\footnotesize{https://github.com/RISC-NYUAD/RCMController}}.

\section{Conclusions}
In this paper, a kinematic control approach for RCM-constrained tool-tip trajectory tracking is developed. The approach is based on a novel task function formulation, which includes a minimal representation of the RCM constraint. Furthermore, the impact of the tool's insertion ratio $\rho$ into the system, as well as, the selection of the starting configuration are discussed. Experimental results showcase the applicability of the approach mainly, but not limited to, RA-MIS operations.   
\section*{Acknowledgements}
This work was partially performed in the Kinesis Lab, Core Technology Platform (CTP) facility of NYUAD.

This work was partially supported by the NYUAD Center for Artificial Intelligence and Robotics (CAIR), funded by Tamkeen under the NYUAD Research Institute Award CG010.
\bibliographystyle{ieeetr}  
\bibliography{conference}

\end{document}